\documentclass[11pt]{article}
\usepackage{acl2015}
\usepackage{times}
\usepackage{url}
\usepackage{latexsym}
\usepackage{graphicx}
\usepackage {amsmath}
\usepackage{multirow}
\usepackage{color,xcolor}
\title{Toxic Comments Hunter : Score Severity of Toxic Comments}

\author{Zhichang Wang$^{*}$ \\
   Southeast University \\
  Nanjing, China \\
  {\tt 213191670@seu.edu.cn} \\\And
  Qipeng Zhu$^{*}$ \\
   Southeast University \\
  Nanjing, China \\
  {\tt 213193319@seu.edu.cn} \\}

\date{}

\begin{document}
\maketitle
\begin{abstract}
The detection and identification of toxic comments are conducive to creating a civilized and harmonious Internet environment. 
In this experiment, we collected various data sets related to toxic comments. 
Because of the characteristics of comment data, we perform data cleaning and feature extraction operations on it from different angles to obtain different toxic comment training sets. 
In terms of model construction, we used the training set to train the models based on TFIDF and finetuned the Bert model separately. 
Finally, we encapsulated the code into software to score toxic comments in real-time.
\end{abstract}

\section{Introduction}

In the comments section of the current major forums, there are mostly friendly comments, but also many toxic comments. These toxic comments are often posted by individuals with low moral standards or those toxic commentators. toxic comments can seriously damage the forum's comment environment and degrade the forum's user experience. At the same time, toxic comments can be confused with negative comments, making it difficult for users to get negative opinions. The identification and scoring of toxic comments using natural language processing technology can effectively identify toxic comments in comments, thus improving the comment environment and reducing the cost of manual review. It is a study of great commercial and moral significance. Therefore, we intend to use the natural language technology, combined with the forum's comment data set, to build a toxic comment scoring model identifying and rating toxic comments on the forum.

In the execution process, we need to build a natural language processing model that can score toxic comments. To achieve this, we first need to find a tagged review dataset. Then we should clean up the comment data according to the rules. Then, we'll use machine learning methods to build the correct natural language model. Finally, we'll deploy the trained model and encapsulate our code into software to detect toxic comments.

To achieve the above goals, we will have to encounter the following challenges:

\begin{itemize}
    \item Complex comment language representations. Different from the composition as well as title, a large number of words are very regular and organized. Comment data sets often have incorrect syntax and exaggerated wording. At the same time, there will be lots of punctuation to enhance the tone. It'll be a big problem for data processing.
    \item The joint training of the datasets. There are some duplicates and wrong labels in our training data sets. Hence, we need to combine the data sets for modeling.
    \item Cross-domain incremental word vectors. We need to consider how to solve the problem of putting the word vector of the subdivision field into a pre-trained vector space, maintaining its clustering and linear properties.

\end{itemize}

Besides the challenges of the task itself, the constraints of our research are mainly time and computing resources. It could take up a month to finish our research, including writing the report. When it comes to computing resources, the highest server standard we can use is a server with 32g memory and a graphics card of 2080Ti, and there is no way to train huge natural language models.

\section{Data Exploration}

While looking for the dataset, we found that a research institute called Jigsaw had posted multiple comment-related datasets on Kaggle. Jigsaw is a research institute affiliated with Google, and its main research direction is to purify the language environment of the Internet, which also includes our research topics. From the datasets published, we selected multiple datasets related to our tasks to assist in our model building. We did some degree of data exploration on these datasets. Here are some of our explorations and the results. Because of the sheer number of datasets, we'll show only some of the core processes here, detailed in the code and experiments attached.

\subsection{Class Dataset}
\begin{figure}[h]
  \centering
  \includegraphics[width=\linewidth]{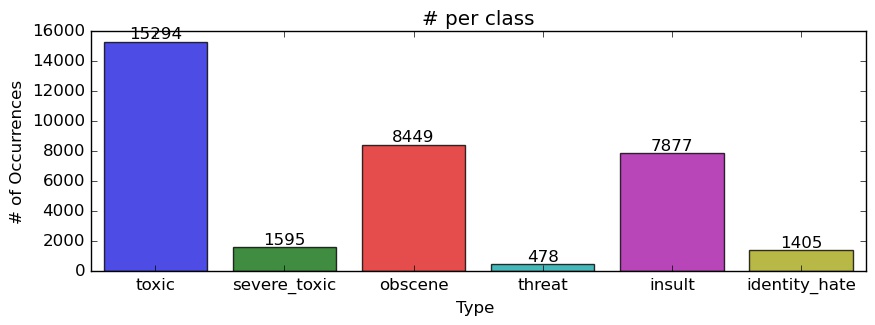}
  \caption{The number of samples with different labels}
  \label{fig:1_class_number}
\end{figure}

The data source is part of Wikipedia's comment data, including the training set and the test set, where the core data table column names are: id, txt, toxic, severetoxic, obscene, threat, insult, identityhate. The first label is ID, the second is the content of the text, and the next few are the results of a label for the toxic text, for a total of five types of toxic labels. The text is 1 if it is toxic and 0 if it is not toxic. Because the dataset is crowdsourced, there are differences in scope between different results. We did some level of data exploration of this dataset. The dataset has 159571 training sets and 153164 test sets. We believe that if there is no toxic evaluation on all indicators, i.e. the indicator sum is 0, we consider the review to be non-toxic. In the training set, toxic comments account for about 10$\%$of the total reviews, and the dataset has a certain imbalance. Categorical classification statistics are performed on different toxic comments on the training set. The statistical results are shown in figure \ref{fig:1_class_number}, where the sample proportion of "toxic" is the highest, the sample of "obscene" and "insult" also has a certain proportion, and the remaining three samples are relatively low.

We counted the number of labels per sample and got the description of the  sample with plenty of tags, and we can see that the number of toxic tags for most comments is not high in toxic comments.

\begin{figure}[h]
  \centering
  \includegraphics[width=\linewidth]{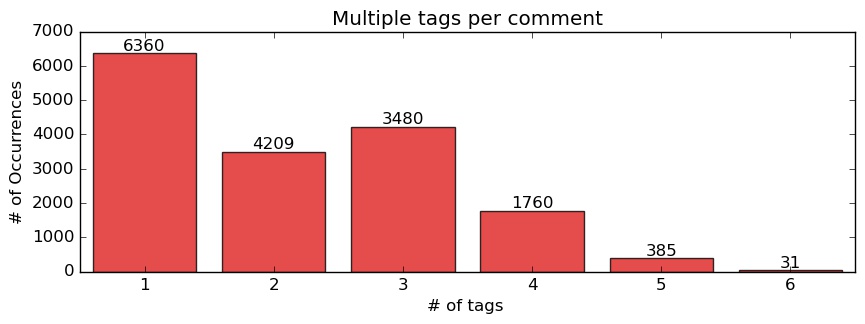}
  \caption{The sample has a number of tags}
\end{figure}

We wanted to explore the correlation between the labels, so we got the chart below, and we could see that the correlation between the indicators was not so high, with the highest correlation being "threat" and "insult" and the higher positive correlation between "obscene". At the same time, all variables are more than 0 between the numbers. Therefore there is no negative correlation.

\begin{figure}[h]
  \centering
  \includegraphics[width=\linewidth]{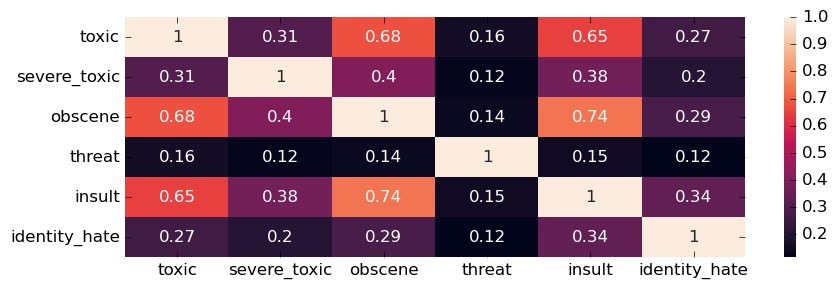}
  \caption{Correlation heat map for different tags}
\end{figure}

We think that the degree of the malice of the comments may be related to the length of the comments. From my understanding, toxic comments are generally vented, so the length of the comments will not be too long. Then we explored the distribution of toxic comments at different comment lengths and drew a violin plot graph. You can see that there are more "clean" comments commented on the distribution as a whole, and the number of statements as a whole will be smaller than the "Toxic" comments.

\begin{figure}[h]
  \centering
  \includegraphics[width=\linewidth]{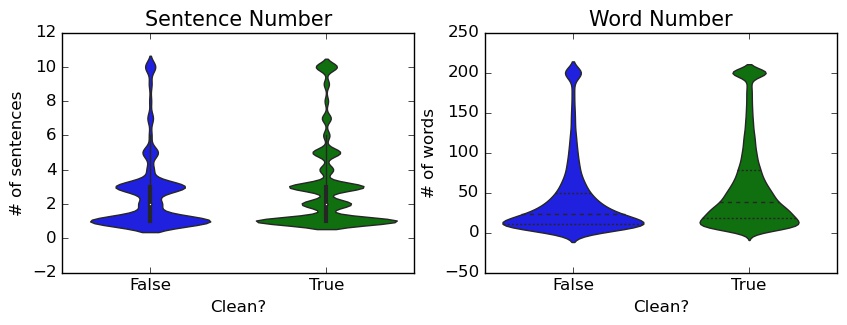}
  \caption{Violinplot of words and sentences}
\end{figure}

\subsection{Bias Dataset}
In the data supplied for this competition, the text of the individual comment is found in the comment-text column. Each comment in Train has a toxicity label. These attributes (and all others) are fractional values that represent the fraction of human raters who believed the attribute applied to the given comment. For evaluation, test set examples with a target of more than 0.5 will be considered in a positive class (toxic).

The data also has several additional toxicity subtype attributes. In the competition, we don't need to predict the subtype, they are included as an additional avenue for research. Subtype attributes are severe toxicity, obscene, threat, insult, identity attack, sexually explicit. There are differences between these indicators and the indicators of the first data set, so they cannot be directly fused when we try to do the dataset fusion.

Additionally, a subset of comments has been labeled with a variety of identity attributes, representing the identities mentioned in the comment.

These metrics can be used to model the deviation of comments, but our goal is to predict the toxins of comments. Therefore, we will not use this part of the dataset. After analyzing the training data set, it is decided to use the 'toxicity-individual-annotations.csv' in the dataset to construct the training data in the end.
\subsection{Multi Dataset}

The data source is part of Wikipedia's review data, including the training and testing set, whose core data sheet is listed as id, comment-text, toxic, severe-toxic, obscene, threat, insult, identity-hate. The first label is the ID, the second is the content of the text, and the next few are a label result of toxic text, with a total of five types of toxic problems. If the text is toxic, the score is not 0. This data table is similar to the previous one, so you can use the same data processing method. At the same time, this dataset provides a sample of the dataset that Bert can learn directly. Here are a few important perspectives to explore.

\begin{figure}[h]
  \centering
  \includegraphics[width=\linewidth]{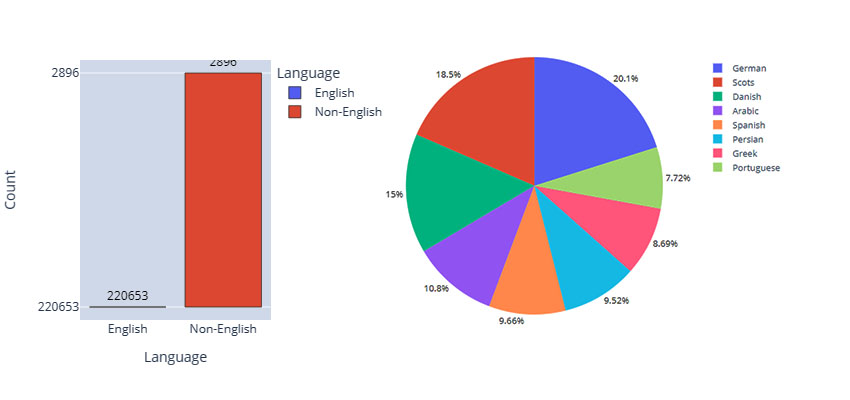}
  \caption{The proportion of different languages}
  \label{fig:3_lan}
\end{figure}

We can see that English comments dominate the training data, with a total of 220636 comments written in English and a mere 2913 comments written in languages other than English. There is a heavy imbalance in the language of comments in the training data. From the figure \ref{fig:3_lan}, we can once again see that German, Danish, and Scots with more than $15\% $of the pie belonging to each of these three languages.

\begin{figure}[h]
  \centering
  \includegraphics[width=\linewidth]{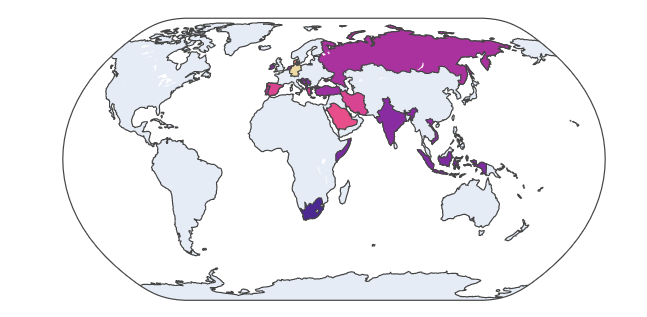}
  \caption{World's distribution of languages}
  \label{fig:3_world}
\end{figure}

From the figure \ref{fig:3_world}, we can see that western Europe and the middle-east are the most represented regions in the dataset. Africa, Asia, and eastern Europe are relatively under-represented.

\begin{figure}[h]
  \centering
  \includegraphics[width=\linewidth]{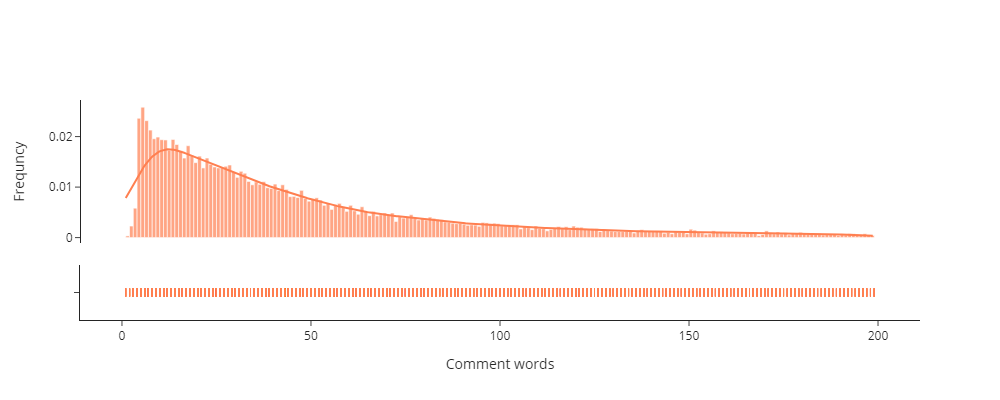}
  \caption{Length distribution of samples}
  \label{fig:3_comment_words}
\end{figure}

From the figure \ref{fig:3_comment_words}, we can see that the distribution of comment words has a strong positive skew with maximum probability density occurring at around 13 words. As the number of words increases over 13, the frequency reduces sharply.

\begin{figure}[h]
  \centering
  \includegraphics[width=\linewidth]{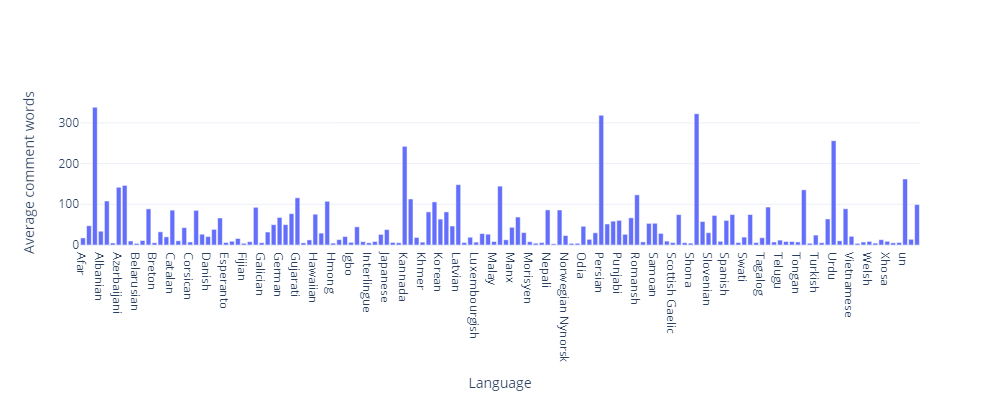}
  \caption{Distribution of comment lengths 
in different languages}
    \label{fig:3_comment_words_language}
\end{figure}

We have plotted the average comment words in each language in figure \ref{fig:3_comment_words_language}. Certain languages tend to have more words on average than other languages. For example, comments written in Akan, Persian, and Sinhala have more than 300 words on average! This may be contributed to the small number of samples in these languages and presence of one or two outliers.

Sentiment and polarity are quantities that reflect the emotion and intention behind a sentence. Now, we will look at the sentiment of the comments using the NLTK (natural language toolkit) library.

\begin{figure}[h]
  \centering
  \includegraphics[width=\linewidth]{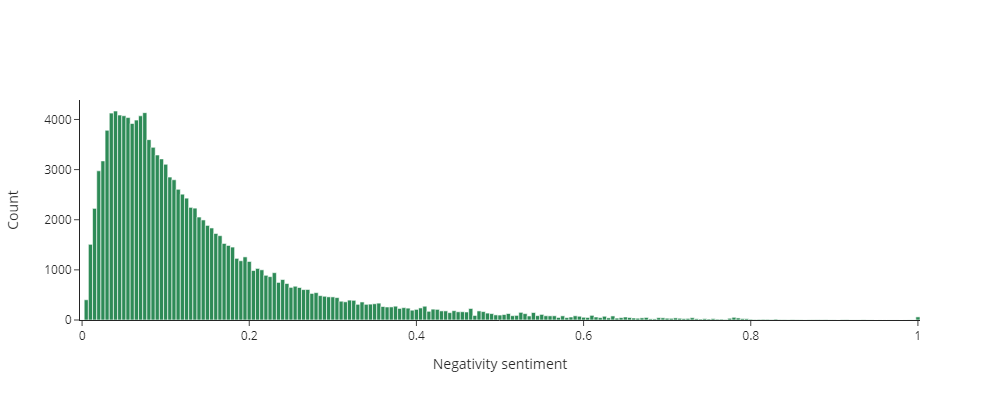}
  \caption{Negative emotional distribution}
  \label{fig:3_neg_sentence}
\end{figure}

As figure \ref{fig:3_neg_sentence} vividly illustrated, we can see that the negative sentiment has a strong rightward (positive) skew, indicating that negativity is usually on the lower side. It suggests that most comments are not toxic or negative. In fact, the most common negative value is around 0.04. Virtually no comments have a negative greater than 0.8.

\begin{figure}[!h]
  \centering
  \includegraphics[width=7.1cm]{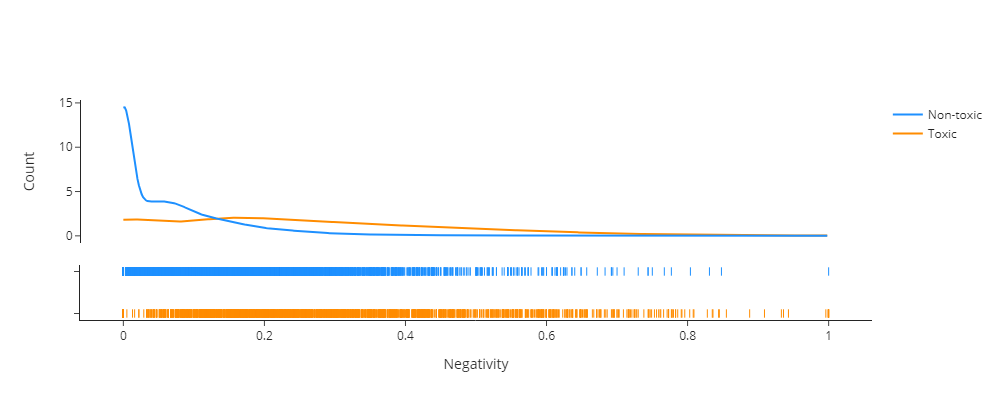}
  \caption{Toxic degree of negative comments}
  \label{fig:3_neg_toxic}
\end{figure}

We have plotted the distribution of negativity for toxic and non-toxic comments as figure \ref{fig:3_neg_toxic} illustrated. We can see that toxic comments have a significantly greater negative sentiment than toxic comments (on average). The probability density of negativity peaks at around 0 for non-toxic comments, while the negativity for toxic comments is minimum at this point. It suggests that comment's very likely to be non-toxic if it has negativity of 0.

\begin{figure}[h]
  \centering
  \includegraphics[width=\linewidth]{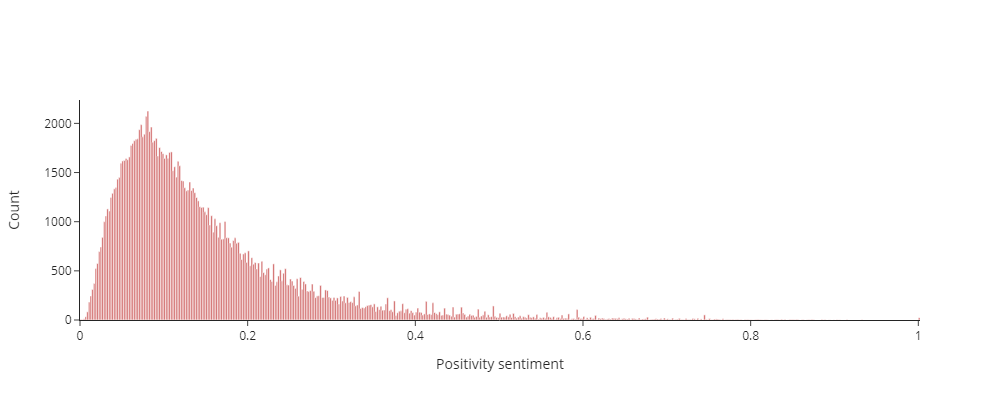}
  \caption{Positive emotional distribution of comments}
  \label{fig:3_pos_sentence}
\end{figure}

From the figure \ref{fig:3_pos_sentence}, we can see that positive sentiment has a strong rightward (positive) skew, indicating that positivity is usually on the lower side. This suggests that most comments do not express positivity explicitly. In fact, the most common negativity value is around 0.08. Virtually no comments have a positivity greater than 0.8.

\begin{figure}[h]
  \centering
  \includegraphics[width=\linewidth]{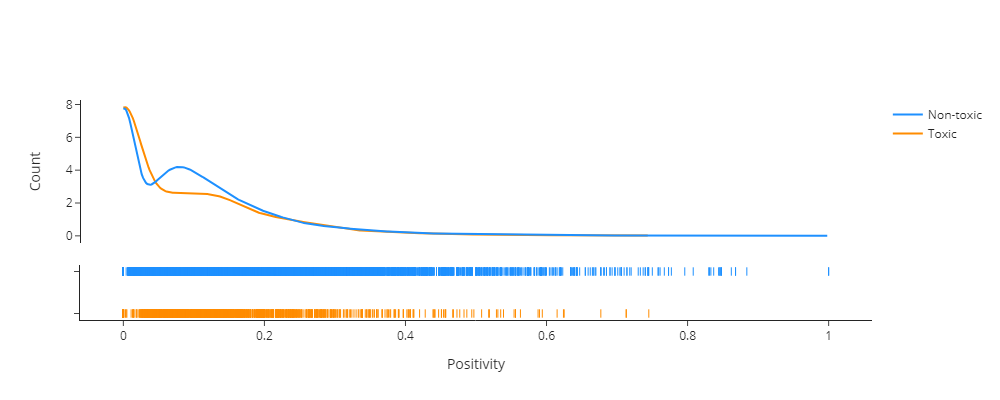}
  \caption{Toxic Degree of positive comments}
  \label{fig:3_pos_toxic}
\end{figure}

We have plotted the distribution of positivity for toxic and non-toxic comments as figure \ref{fig:3_pos_toxic} illustrated. We can see that both the distributions are very similar, indicating that positivity is not an accurate indicator of toxicity in comments. 

\begin{figure}[h]
  \centering
  \includegraphics[width=6.3cm]{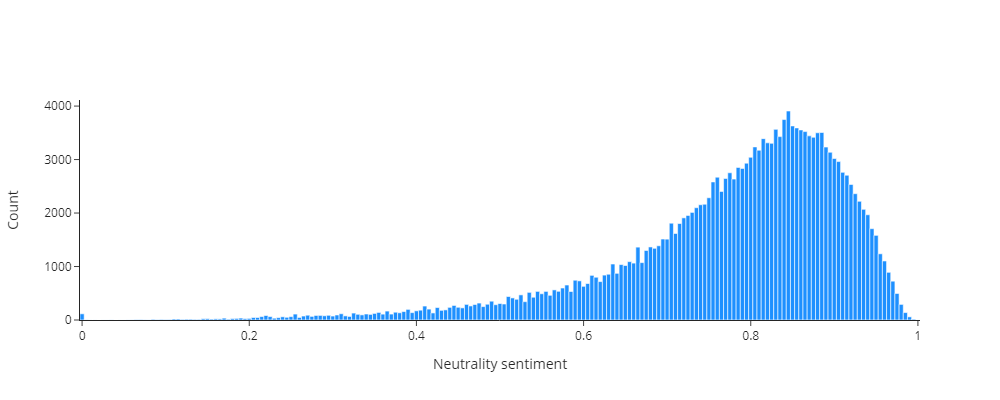}
  \caption{Neutrality distribution of the comments}
  \label{fig:3_neu_sentence}
\end{figure}

From the figure \ref{fig:3_neu_sentence}, we can see that the neutrality sentiment distribution has a strong leftward (negative) skew, which is in contrast to the negativity and positivity sentiment distributions. This indicates that the comments tend to be very neutral and unbiased in general. This also suggests that most comments are not highly opinionated and polarizing, meaning most comments are non-toxic.

\begin{table*}[]

\setlength{\tabcolsep}{2.2mm}{
\renewcommand\arraystretch{1.0}
\begin{tabular}{ll|ll|ll|ll|ll}
\hline
                                                       &                   & \multicolumn{2}{l|}{\textbf{Class Dataset}}     & \multicolumn{2}{l|}{\textbf{Bias Dataset}} & \multicolumn{2}{l|}{\textbf{Multi Dataset}} & \multicolumn{2}{l}{\textbf{Ruddit Dataset}} \\ \cline{3-10} 
                                                       &                   & \textbf{clean0}               & \textbf{clean1} & \textbf{clean0}      & \textbf{clean1}     & \textbf{clean0}      & \textbf{clean1}      & \textbf{clean0}      & \textbf{clean1}      \\ \hline
\multicolumn{1}{l|}{}                                  & \textbf{LightGBM} & 0.5917                        & 0.6365          & 0.5498               & 0.5363              & 0.644                & 0.6365               & 0.5868               & 0.5872               \\
\multicolumn{1}{l|}{}                                  & \textbf{MLP}      & 0.6307                        & 0.6496          & 0.6246               & 0.6231              & 0.6537               & 0.655                & 0.628                & 0.617                \\
\multicolumn{1}{l|}{}                                  & \textbf{Ridge}    & {\color[HTML]{CB0000} 0.6722} & 0.6708          & 0.6382               & 0.6403              & 0.6717               & 0.6708               & 0.6324               & 0.6303               \\
\multicolumn{1}{l|}{\multirow{-4}{*}{\textbf{Tfidf0}}} & \textbf{SVM}      & 0.6413                        & 0.6568          & 0.6325               & 0.6343              & 0.6607               & 0.6568               & 0.6292               & 0.6299               \\ \hline
\multicolumn{1}{l|}{}                                  & \textbf{LightGBM} & 0.5921                        & 0.6215          & 0.5512               & 0.5439              & 0.6228               & 0.6215               & 0.5808               & 0.5809               \\
\multicolumn{1}{l|}{}                                  & \textbf{MLP}      & 0.6257                        & 0.6496          & 0.582                & 0.5806              & 0.6416               & 0.6445               & 0.6134               & 0.6124               \\
\multicolumn{1}{l|}{}                                  & \textbf{Ridge}    & 0.6357                        & 0.6542          & 0.5968               & 0.5969              & 0.652                & 0.6542               & 0.6107               & 0.6109               \\
\multicolumn{1}{l|}{\multirow{-4}{*}{\textbf{Tfidf1}}} & \textbf{SVM}      & 0.6279                        & 0.6493          & 0.5858               & 0.5863              & 0.6453               & 0.6493               & 0.6128               & 0.617                \\ \hline
\multicolumn{1}{l|}{}                                  & \textbf{LightGBM} & 0.6168                        & 0.6157          & 0.4866               & 0.5014              & 0.6168               & 0.6157               & 0.5828               & 0.5829               \\
\multicolumn{1}{l|}{}                                  & \textbf{MLP}      & 0.639                         & 0.6424          & 0.5703               & 0.5786              & 0.6484               & 0.6459               & 0.6102               & 0.608                \\
\multicolumn{1}{l|}{}                                  & \textbf{Ridge}    & 0.6351                        & 0.6629          & 0.5735               & 0.5731              & 0.6631               & 0.6629               & 0.6189               & 0.6194               \\
\multicolumn{1}{l|}{\multirow{-4}{*}{\textbf{Tfidf2}}} & \textbf{SVM}      & 0.654                         & 0.6533          & 0.5581               & 0.5587              & 0.654                & 0.6533               & 0.6156               & 0.6158               \\ \hline
\multicolumn{1}{l|}{}                                  & \textbf{LightGBM} & 0.6147                        & 0.6144          & 0.5002               & 0.4698              & 0.6147               & 0.6144               & 0.5816               & 0.5825               \\
\multicolumn{1}{l|}{}                                  & \textbf{MLP}      & 0.6489                        & 0.6408          & 0.5376               & 0.5586              & 0.6477               & 0.6424               & 0.6006               & 0.597                \\
\multicolumn{1}{l|}{}                                  & \textbf{Ridge}    & 0.6629                        & 0.6637          & 0.5743               & 0.5734              & 0.6629               & 0.6637               & 0.6183               & 0.6185               \\
\multicolumn{1}{l|}{\multirow{-4}{*}{\textbf{Tfidf3}}} & \textbf{SVM}      & 0.6551                        & 0.6553         & 0.5584               & 0.5573              & 0.6551               & 0.6553               & 0.6153               & 0.6148               \\ \hline 

\end{tabular}
}
\caption{Experimental results of different machine learning models on the validation set}
\end{table*}

\begin{figure}[h]
  \centering
  \includegraphics[width=\linewidth]{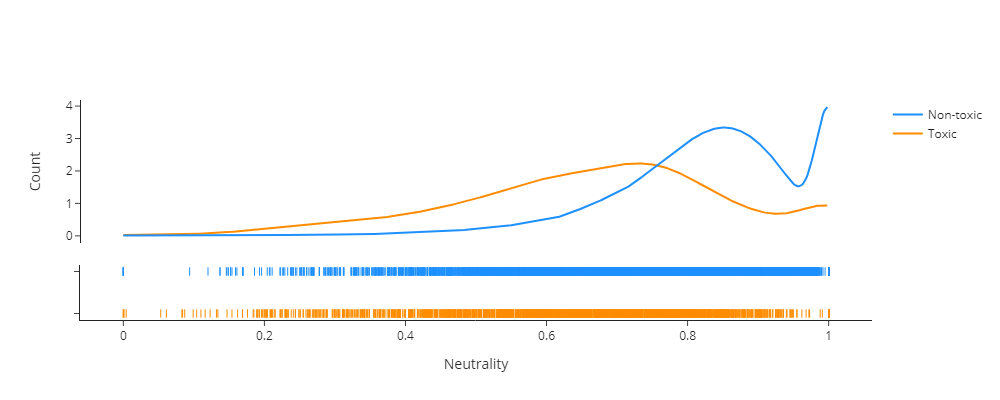}
  \caption{Degree of toxic of neutral comments}
  \label{fig:3_neu_toxic}
\end{figure}

We can see that non-toxic comments tend to have a higher neutrality value than toxic comments on average from figure \ref{fig:3_neu_toxic}. The probability density of the non-toxic distribution experiences a sudden jump at 1, and the probability density of the toxic distribution is significantly lower at the same point. It suggests that comment with neutrality close to 1 is more likely to be non-toxic than toxic.

\begin{figure}[h]
  \centering
  \includegraphics[width=\linewidth]{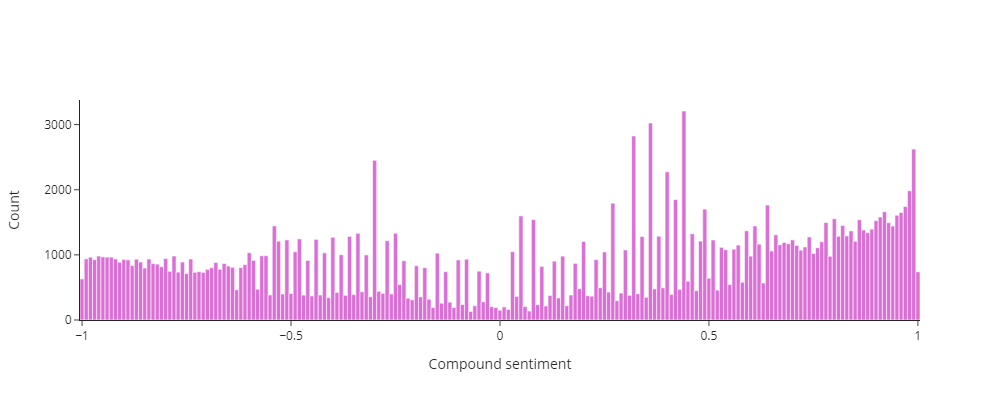}
  \caption{Compound emotional distribution of the comments}
  \label{fig:3_all_sentence}
\end{figure}

As figure \ref{fig:3_all_sentence} shown, we can also see that compound sentiment is evenly distributed across the spectrum (from -1 to 1) with very high variance and random peaks throughout the range.

\begin{figure}[h]
  \centering
  \includegraphics[width=\linewidth]{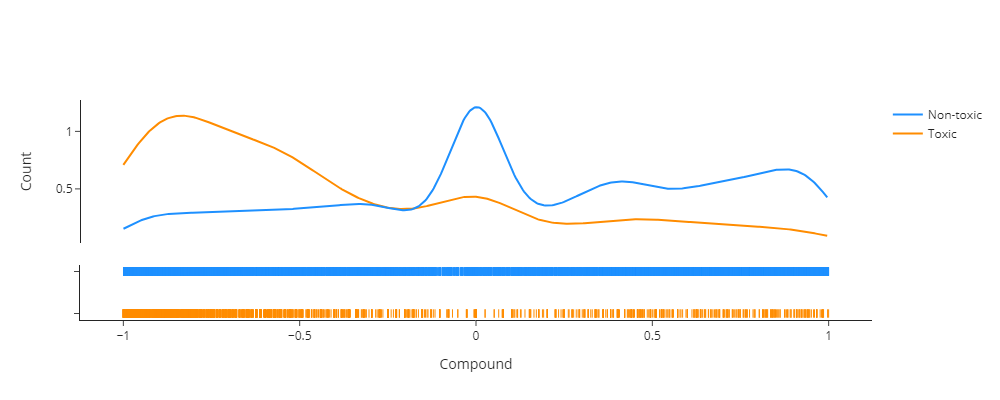}
  \caption{Toxic degree of compound comments}
  \label{fig:3_all_toxic}
\end{figure}

From the next picture, figure \ref{fig:3_all_toxic}, we can see that compound sentiment tends to be higher for non-toxic comments compared to toxic comments. The non-toxic distribution has a leftward (negative) skew, while the toxic distribution has a positive (rightward) skew. It indicates that non-toxic comments tend to have a higher compound sentiment than toxic comments on average.

\subsection{Validation Dataset}
The data source is some of Wikipedia's comment data, with no training set, where the core data table is listed as worker, less-toxic, more-toxic. The first label is the labeler's ID, and the next two labels are text. The labeler determines the degree of the malice of both texts. In subsequent experiments, we used this dataset as a validation set.

\subsection{Ruddit Dataset}
The data source is part of Reddit's English review data, where the core data sheets are: post-id, comment-id, txt, URL, offensiveness-score. Post-id is the ID of the post. comment-id is the ID of the comment. The txt is the text content. Offensiveness-score is the rating of the office. The range of offensiveness-score is from -1 to 1.


\section{Model Construction}
When we build the model, we are based on the theoretical assumption that toxic comments will have similarities in semantic and linguistic expressions. In the process of EDA, we found that words such as *Fxxk, SHxx* and other words with toxic labels are likely to be high, which is a semantic similarity. Simultaneous image with *!!!!! *, as well as reviews that use abbreviations a lot are also highly likely to be toxic comments. For the characteristics of the comment data, we use two different methods to model, namely the feature extraction modeling method based on TFIDF and the Fine-tuning method based on Bert's pre-trained model.
\subsection{Model Based On TFIDF}
\subsubsection{Data Cleaning}

We mainly use regular expressions to reconstruct unreasonable natural language. TFIDF requires feature extraction based on the word frequency of words, but there are all kinds of strange word expressions in the review data. For example: $sh***t, FXXxk\ uuu$. These languages expand the range of words and are not easily captured during feature extraction. Therefore, we have designed two sets of natural language cleaning templates $clean_0, clean_1$. $clean_0$ is based on a regular natural language task, the main elimination is case, duplicate words. $clean_1$ for datasets, multi-language, multi-expression, multi-reference features. Use regular expressions to remove emoticons in comments, words that are not in English, and some IP and web page identifiers. 

\subsubsection{Feature extraction}

The feature extraction part we use is the TFIDF method. TFIDF is a common weighting technique for information retrieval and text mining. TFIDF is a statistical method used to assess the importance of a word to a file set or one of the documents in a corpus. The importance of a word increases proportionally with the number of times it appears in a file, but decreases inversely with the frequency it appears in the corpus. Where there is the calculation method of tf:

\begin{equation}
\begin{aligned}
TF(t, d)=\frac{f_{t, d}}{\sum_{t^{\prime} \in d} f_{t^{\prime}, d}}
\end{aligned}
\end{equation}

where $f_{t,d}$ is the raw count of a term in a document, i.e., the number of times that term t occurs in a document  There are various other ways to define the term. We calculate idf as equation \ref{eq:IDF}:

\begin{equation}
\begin{aligned}
IDF(t, D)=\log \frac{N}{|\{d \in D: t \in d\}|}
\end{aligned}
\label{eq:IDF}
\end{equation}

$N$ is the total number of documents in the corpus $N=|D|$.
$|\{d \in D: t \in d\}|$is the number of documents where the term $t$ appears (i.e., $TF(t, d) \neq 0$). If the term is not in the corpus, this will lead to a division-by-zero. So it is common to adjust the denominator. During the process of the experiment, we limited the initial TFIDF method and obtained the feature only if it appeared more than 3 times in the corpus and the frequency of occurrence was less than 0.5. To widely test the results, we designed four different TFIDF treatments: $tfidf_0$, $tfidf_1$, $tfidf_2$, $tfidf_3$ . For the first two of these, we use character-based gram extraction methods. In detail, we select the number of characters for 3-5 to extract features. The second method we expect to look at is the effect of features on the results by reducing the number of features. In the third and fourth scenarios are that we use word-based feature extraction methods, the former we only extract 1-2 grams of word expression characteristics, the latter we extract 1-5 grams of word expression characteristics. We want to explore the impact of different grams on the results.

\subsubsection{Model  And Evaluation Indicator}

On the model, we modeled four common machine learning algorithms: Ridge Regression, SVM Regression, MLP Regression, lightGBM. It is worth mentioning that we choose the linear nuclear model for SVM. In the selection of indicators, we validate the effect of the model based on the fourth data set. If the predicted results are consistent with the results of the fourth dataset, we believe that the predicted results are correct. The formula of the indicator is shown below, where $y$ represents the predicted result, $y^0_p$ represents the predicted outcome of the less malicious comment on the corpus $p$, and $y^1_p$ represents the predicted result of the more malicious comments on the corpus. $D_{LEN}$ represents the length of the validation set.

\begin{equation}
\begin{aligned}
ACC=\frac{\sum_{p\in D}\lfloor y^0_p-y^1_p+1 \rfloor}{D_{LEN}}
\end{aligned}
\end{equation}

\subsubsection{Experiment}

Comparing the different models, you can see that Ridge Regression performs well on most regression tasks. LightGBM's \cite{ke2017lightgbm} predictions are the most backward. On some tasks, the SVM can achieve the same performance as Ridge Regression. The possible reason is that the extracted features are sparse and less characterized, so the traditional model will perform better. 

From the perspective of datasets, Class Dataset and Multi Dataset have the best performance in this task. We think the possible reason for the model'spoor performance in bias dataset is that the distribution of bias Dataset is inconsistent with the validation set. On the other hand, the poor performance in Ruddit Dataset should be caused by its too small collective amount of data. 

Looking at the different cleaning methods, we can see that the distinction between the two is smaller, some tasks are good on clean1, and some think that clean0 is good. This shows that both cleaning methods are effective, and we consider mixing them in subsequent reasoning. 

For different TFIDF methods, we can see that the first method still works best. A possible reason, we think, is that it is more suitable for this more confusing set of comments. We will also use this method in subsequent inferring. 

What was not shown in the experiment was that there was a big gap in the training time of different tasks. In general, the training time of Ridge Regression, LightGBM\cite{ke2017lightgbm} was within our acceptable range. The remaining two models take much longer to train than the two models.

\subsection{Model Based On Bert}

We tried to finish our mission using the Bert \cite{kenton2019bert} model based on the pre-trained model. We model each piece of text as a vector of 128 dimensions in length for model training. We built a linear layer at the last layer of the model for Fine-tuning. We use Margin Ranking Loss as our loss function. For batch data $D (x_1, x_2, y)$ containing $N$ samples, $x_1$, $x_2$ is the two sorted inputs, $y$ represents the real label and belongs to the $\{1,-1\}$. When $y$ is $1$ , $x_1$ should be before $x_2$. When $y$ is $-1$, $x_1$ should be after $x_2$. The loss for the $n$ sample is calculated as follows:

\begin{equation}
\begin{aligned}
l_{n}=\max (0,-y *(x 1-x 2)+\operatorname{margin})
\end{aligned}
\end{equation}

If $x_1, x_2$ is sorted correctly and $-y(x_1-x_2)>\text{MARGIN}$, else is $(-y *(x_1-x_2)+\operatorname{margin)}$. The validation indicator of the final model can be similar to that of the previous method. But the model using Bert has the problem of large model space and slow inferring speed. Moreover, it's difficult to optimize it due to the limitations of our strength. 
\section{Application}

On the deployment of the model, we select the model using tfidf and Ridge Regression based on the Class Dataset. Under these conditions, the model has a good performance in the evaluation index. And the speed of inferring is within the acceptable range. Last but not least, we encapsulated the code into software where users can input whatever comments they want to input to find out how toxic the comments are  in real-time. 

\section{Summary}
After this experiment, we cleverly applied the knowledge learned in the natural language course. From a simple machine learning model based on tfidf to the popular bert model, we are constantly exploring how to achieve better experimental results. Although due to time and equipment limitations, we still benefit a lot. This mission is that we have a more comprehensive understanding of the NLP mission. Especially when we face the complex environment in reality, we have to consider more than just how to build a model. How to clean the data and accurately extract the characteristics of the data will be an important basis for completing the task. In the winter vacation, we will try more complex NLP tasks, such as dialogue robots. I believe we will go further and further on the road of NLP! That's all, thank you!

\bibliographystyle{acl}
\bibliography{ref}

\begin{thebibliography}{}

\bibitem[\protect\citename{Ke \bgroup et al.\egroup }2017]{ke2017lightgbm}
Guolin Ke, Qi~Meng, Thomas Finley, Taifeng Wang, Wei Chen, Weidong Ma, Qiwei
  Ye, and Tie-Yan Liu.
\newblock 2017.
\newblock Lightgbm: A highly efficient gradient boosting decision tree.
\newblock {\em Advances in neural information processing systems},
  30:3146--3154.

\bibitem[\protect\citename{Kenton and Toutanova}2019]{kenton2019bert}
Jacob Devlin Ming-Wei~Chang Kenton and Lee~Kristina Toutanova.
\newblock 2019.
\newblock Bert: Pre-training of deep bidirectional transformers for language
  understanding.
\newblock In {\em Proceedings of NAACL-HLT}, pages 4171--4186.

\end{thebibliography}

\end{document}